\documentclass{opendatalab}
\usepackage{algorithm}
\usepackage{algpseudocode}
\usepackage[utf8]{inputenc}
\usepackage{xcolor}
\usepackage[most]{tcolorbox}
\usepackage{longtable}
\usepackage{listings}
\usepackage[numbers]{natbib}
\usepackage{enumitem}
\usepackage{graphicx}
\usepackage{xspace} 
\usepackage{hyperref}
\usepackage{xspace}
\usepackage[table]{xcolor}
\definecolor{baselinegray}{HTML}{EEF2F7}

\makeatletter
\DeclareRobustCommand\onedot{\futurelet\@let@token\@onedot}
\def\@onedot{\ifx\@let@token.\else.\null\fi\xspace}

\makeatother
\definecolor{codegreen}{rgb}{0,0.6,0}
\definecolor{codegray}{rgb}{0.5,0.5,0.5}
\definecolor{codepurple}{rgb}{0.58,0,0.82}
\definecolor{backcolour}{rgb}{0.95,0.95,0.92}
\definecolor{promptcolor}{HTML}{D1D0F2}
\definecolor{promptcolorheader}{HTML}{bdbcec}

\definecolor{promptcolor}{HTML}{D1D0F2}
\definecolor{promptcolorheader}{HTML}{bdbcec}

\newtcolorbox{promptbox}[1][]{
  enhanced, breakable,
  top=0.3em,bottom=0.3em,left=0.5em,right=0.5em,
  toptitle=0.3em,bottomtitle=0.2em,boxsep=0pt,
  colframe=promptcolorheader, colback=promptcolor!50, boxrule=0.5pt,
  width=\columnwidth, 
  title={\footnotesize #1} 
}
\lstdefinestyle{promptstyle}{
    backgroundcolor=\color{backcolour},   
    commentstyle=\color{codegreen},
    keywordstyle=\color{magenta},
    numberstyle=\tiny\color{codegray},
    stringstyle=\color{codepurple},
    basicstyle=\ttfamily\footnotesize,
    breakatwhitespace=false,         
    breaklines=true,                 
    captionpos=b,                    
    keepspaces=true,                 
    numbers=left,                    
    numbersep=5pt,                  
    showspaces=false,                
    showstringspaces=false,
    showtabs=false,                  
    tabsize=2
}
\title{Better Starts, Better Ends: Bootstrapped Iterative Self-Reasoning Distillation for Compressed Reasoning}

\author{
    \parbox{\linewidth}{\raggedright
        Leichao Dong$^{1}$ \quad
        Dongxu Zhang$^{1}$ \quad
        Yiding Sun$^{1}$ \quad
        Qirui Wang$^{1}$ \quad
        Yuhan Wang$^{2}$ \\ \vspace{0.15cm}
        Lin Chen$^{1}$ \quad
        Jihua Zhu$^{1}$ \\ \vspace{0.15cm}
    }
}

\affiliation{
    \vspace{0.2cm} 
    \parbox{\linewidth}{\raggedright \small
        $^1$Xi'an Jiaotong University \quad
        $^2$Peking University \quad
    }
}

\abstract{
Large reasoning models often solve problems through long chain-of-thought (CoT) traces, yet much of this computation is spent on redundant derivations, repeated self-verification, and detours that do not improve the final answer. Existing on-policy self-distillation methods reduce this cost by matching a student model to a concise copy of itself on prefixes sampled from the student's own rollouts. We show that this objective has an initialization bottleneck. Since supervision is applied only to visited prefixes, training from a verbose base model places the KL loss on contexts that are often noisy, redundant, or already off track. In such regions, a concise teacher can provide only local corrections, while the student continues to explore trajectories that an efficient reasoner should avoid. In this paper, we propose \textsc{BIRD} (\textbf{B}ootstrapped \textbf{I}terative Self-\textbf{R}easoning \textbf{D}istillation), a two-stage self-reasoning distillation method that improves the rollout distribution before on-policy training. BIRD first samples concise solutions from the base model under a brevity instruction, keeps only answer-correct traces, and performs a lightweight prompt-switch SFT step. The traces are generated with the brevity instruction but learned under the original task prompt, turning instruction-induced conciseness into a default reasoning behavior. Starting from this warm model, BIRD then applies on-policy reverse-KL distillation with a concise self-teacher, now on cleaner and more informative prefixes. Across Qwen3 series models, BIRD achieves a stronger accuracy-efficiency trade-off than prompting and cold-start on-policy distillation on MATH-500 and AIME benchmarks. On Qwen3-8B, it improves MATH-500 accuracy from 86.2\% to 92.0\% while reducing the average response length from 3,099 to 1,115 tokens. These results highlight prefix support as a central factor in efficient reasoning distillation.

}

\metadata[Contact]{Leichao Dong, \email{leichaodong@stu.xjtu.edu.cn}}
\metadata[Code]{\url{https://github.com/KawhiC/BIRD}}

\begin{document}

\maketitle


\begin{figure*}[t]
\centering
\includegraphics[width=\textwidth]{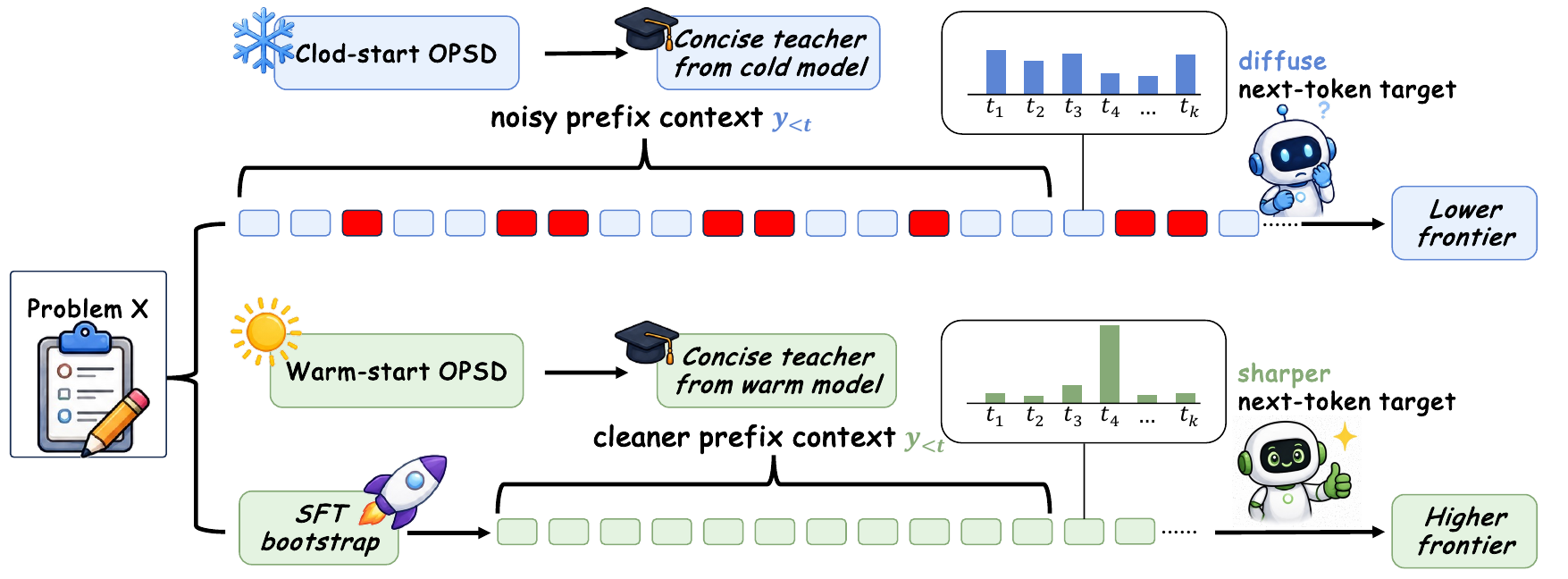}
\caption{Prefix support determines where on-policy KL supervision is applied. Cold-start OPSD and warm-start OPSD use the same concise-teacher construction, but they induce different student rollouts. Starting from the base model exposes the KL loss to noisy or off-track prefixes, where the next-token target can be diffuse and less useful for compression. \textsc{BIRD} first applies SFT bootstrapping and shifts the visited prefixes toward cleaner reasoning contexts. The same prefix-local KL objective is then evaluated on sharper teacher targets and reaches a higher compression frontier.}
\label{fig:intro}
\end{figure*}

\section{Introduction}
\label{sec:int}
Chain-of-thought (CoT) reasoning has become a central capability of large language models, enabling complex problems to be solved through explicit intermediate steps \cite{wei2022chain,huang2026cfms}. Yet this capability also makes inference substantially more expensive. Reasoning traces often span thousands of tokens and contain substantial redundancy, including repeated derivations, revisited conclusions, and excessive self-verification\cite{peng2025revisiting,li2026making,chiang2024over,zhang2026llms,wei2025stop}. In many cases, this additional computation contributes little to the final answer, making reasoning length a major obstacle to efficient deployment. Reasoning compression addresses this problem by shortening generated traces while preserving, and ideally improving, task accuracy.

Recent work has explored several routes toward this goal. Training-free methods control the reasoning process at inference time through concise prompting, early stopping, or self-correction \cite{renze2024benefits,yang2025dynamic,wu2026intern}. RL-based methods introduce length-sensitive rewards to discourage unnecessary tokens\cite{aggarwal2025l1,wan2026mitigating,li2026stepwise,li2026leash}. SFT-based methods imitate compressed traces collected from static datasets, often using external models as trace generators \cite{xia2025tokenskip,zhang2026chain,ma2025cot,fan2026ctrlcot}. From a distribution-matching perspective, these methods can be viewed as different ways of shifting the model toward a more concise reasoning distribution \cite{zhang2026llms,wei2025stop,zhang2026pointcot}. On-policy self-distillation (OPSD) provides a self-contained alternative, in which the model itself acts as a dynamic concise teacher when conditioned on a conciseness instruction \cite{zhao2026self,sang2026crisp}. This removes the need for an external teacher during distillation and suggests that concise reasoning behavior is already latent in the base model.

A key limitation, however, lies in the support on which OPSD applies its supervision. The objective is prefix-local: at each update, the student first samples a rollout, and the KL loss is evaluated only on prefixes $y_{<t}$ visited by that rollout \cite{agarwal2024policy,zhang2026opsdl}. The concise teacher is therefore not queried over an abstract space of high-quality solutions, but on the concrete partial trajectories produced by the current student. When training starts from a verbose base model, these early trajectories are often noisy, redundant, and unreliable \cite{sui2025stop,zhao2026rosd}. Once a prefix already contains a detour, an unnecessary self-check, or an incorrect intermediate step, the teacher must condition on that degraded context. The resulting signal can correct the next token locally, but it remains anchored to trajectory regions that an efficient reasoner should avoid.

This creates a cold-start prefix-support bottleneck. Cold-start OPSD and warm-start OPSD may use the same concise-teacher construction, but they expose the teacher to different prefix distributions. As illustrated in Figure~\ref{fig:intro}, cold-start training places the KL objective on prefixes inherited from the base model's verbose rollouts. A better initialization changes the student's rollout support before online distillation begins, allowing the same KL objective to operate on shorter, more reliable, and more informative reasoning contexts. The central issue is therefore not only how to construct a concise teacher, but also how to make the student visit prefixes where that teacher can provide actionable compression signal. \textsc{BIRD} (Bootstrapped Iterative Self-Reasoning Distillation) is designed around this observation. The method keeps the concise self-teacher unchanged and instead reshapes the rollout distribution on which the teacher is queried. It first performs a lightweight LoRA-based offline bootstrap on correctness-filtered self-generated concise traces \cite{hu2022lora}. These traces are sampled from the base model under a conciseness instruction, but learned under the original task prompt. This prompt switch converts instruction-induced conciseness into a default reasoning behavior rather than an inference-time artifact. Starting from this warm model, OPSD concentrates prefix-local KL updates on cleaner reasoning contexts instead of noisy or off-track prefixes produced by cold-start rollouts.

Experiments on Qwen3 series models \cite{yang2025qwen3} and DeepSeek-R1-Distill-Llama \cite{guo2025deepseek} across MATH-500, AIME 2024, and AIME 2025 show consistent improvements in the accuracy-efficiency trade-off. Compared with prompting and cold-start OPSD, \textsc{BIRD} produces shorter traces while preserving or improving accuracy. Training dynamics further show that cold-start OPSD converges to a lower compression frontier and does not close the gap with longer optimization. The gains become stronger with model scale and transfer across model families, highlighting prefix support as a central factor in on-policy reasoning compression. The contributions are summarized as follows:
\begin{itemize}
\item To the best of our knowledge, this work is the first to identify the cold-start prefix-support bottleneck as a key failure mode in on-policy reasoning compression, where prefix-local KL supervision is applied to verbose, redundant, or off-track student trajectories.

\item We formulate a support-placement principle for OPSD: the effectiveness of a concise self-teacher depends not only on the target distribution it defines, but also on the rollout prefixes where this target is queried.

\item We introduce \textsc{BIRD}, a bootstrapped iterative reasoning distillation method that uses correctness-filtered concise traces to warm-start the student, then performs OPSD on cleaner and more informative prefix support.

\item Extensive experiments on Qwen3 and DeepSeek-R1-Distill-Llama across MATH-500, AIME 2024, and AIME 2025 show that \textsc{BIRD} consistently improves the accuracy-efficiency frontier over prompting and cold-start OPSD.
\end{itemize}

\section{Related Work}
\subsection{RL with length penalties.}
RL-based methods encourage compression by reshaping the reward landscape. L1 \cite{aggarwal2025l1} uses GRPO to condition reasoning models on target lengths, enabling controllable reasoning depth. DiPO \cite{wan2026mitigating} derives difficulty signals from self-generated reasoning statistics and uses them to modulate length penalties. SwAP \cite{li2026stepwise} applies step-level adaptive penalization based on on-policy log-probability improvement, concentrating penalties on low-importance reasoning steps. Leash \cite{li2026leash} adaptively adjusts length penalties so that large reasoning models can shorten outputs while preserving accuracy. These methods directly optimize the accuracy-efficiency trade-off, but their effectiveness depends on reward design and often requires careful tuning of penalty strength to balance correctness and brevity.

\subsection{SFT on compressed chains-of-thought.}
SFT-based methods train on compressed reasoning traces collected~\cite{tang2026mitigatinghallucinationsinterlayerconsistency,tang2026seememitigatinghallucinationslarge,guo2026fademitigatinghallucinationsreducing,wang2026pointrft} or constructed offline~\cite{zhang2026pointcot,zhang2025not,zhang2026not}. TokenSkip \cite{xia2025tokenskip} learns to identify and skip low-importance reasoning tokens. V-Skip \cite{zhang2026chain} uses a visual-anchored information bottleneck for selective token pruning. CoT-Valve \cite{ma2025cot} identifies a parameter-space direction that controls reasoning length, and CtrlCoT \cite{fan2026ctrlcot} combines semantic abstraction with logic-preserving pruning. The main advantage of this family is its simplicity, as the desired compressed behavior is explicitly specified by the training data. Its limitation is potential distribution mismatch: the student is trained on static traces that may differ from the trajectories it later generates at inference time.

\subsection{On-policy self-distillation.}
OPSD \cite{zhao2026self} conditions the teacher on ground-truth answers and improves sample efficiency over RL. TIP \cite{xu2026tip} further studies on-policy distillation from a token-importance perspective, assigning greater weight to more informative tokens during training. FiRe-OPD \cite{li2026filter} rethinks the granularity of OPD optimization by first filtering low-quality trajectories and then softly reweighting tokens within the retained trajectories according to their learning value. CRISP \cite{sang2026crisp} shows that a conciseness instruction alone can serve as the teacher context, removing the need for ground-truth answers during distillation. However, existing OPSD methods mainly focus on teacher construction or token weighting, while paying less attention to the prefix support on which distillation is actually applied.

\section{Method}

\subsection{Preliminaries and Distribution-Matching View}

A language model $\pi_\theta$ with parameters $\theta$ defines a conditional distribution over token sequences. Given an input prompt $x$, the model generates an output sequence $y=(y_1,y_2,\dots,y_T)$ autoregressively. At decoding step $t$, the next-token distribution is conditioned on both the original prompt and the previously generated prefix $y_{<t}$, which represents the current partial reasoning state. The probability of the complete response factorizes as:
\begin{equation}
\pi_\theta(y \mid x) = \prod_{t=1}^{T} \pi_\theta(y_t \mid x, y_{<t}),
\label{eq:ar}
\end{equation}
where $y_{<t} = (y_1, \dots, y_{t-1})$ denotes the prefix generated up to position $t{-}1$. This prefix-conditioned view is particularly important for reasoning models with extended generation traces. Intermediate tokens are not merely surface realizations, but determine the context from which subsequent reasoning steps and the final answer are generated.

In reasoning tasks, $y$ typically consists of a reasoning trace $r$ followed by a final answer $a$. Reasoning compression can therefore be viewed as modifying $\pi_\theta(y\mid x)$ to favor shorter and more direct trajectories while preserving correctness. We use distribution matching as a unifying formal lens. Let $\pi_{\mathrm{target}}$ denote a concise target distribution induced by inference-time controls, reward shaping, offline traces, or a dynamic self-teacher. Compression can be written as:
\begin{equation}
\min_\theta\; D\bigl(\pi_\theta, \pi_{\mathrm{target}}\bigr),
\label{eq:dist_match}
\end{equation}
where methods differ not only in how $\pi_{\mathrm{target}}$ is constructed, but also in where the matching loss is evaluated.

\begin{figure*}[t]
\centering
\includegraphics[width=\textwidth]{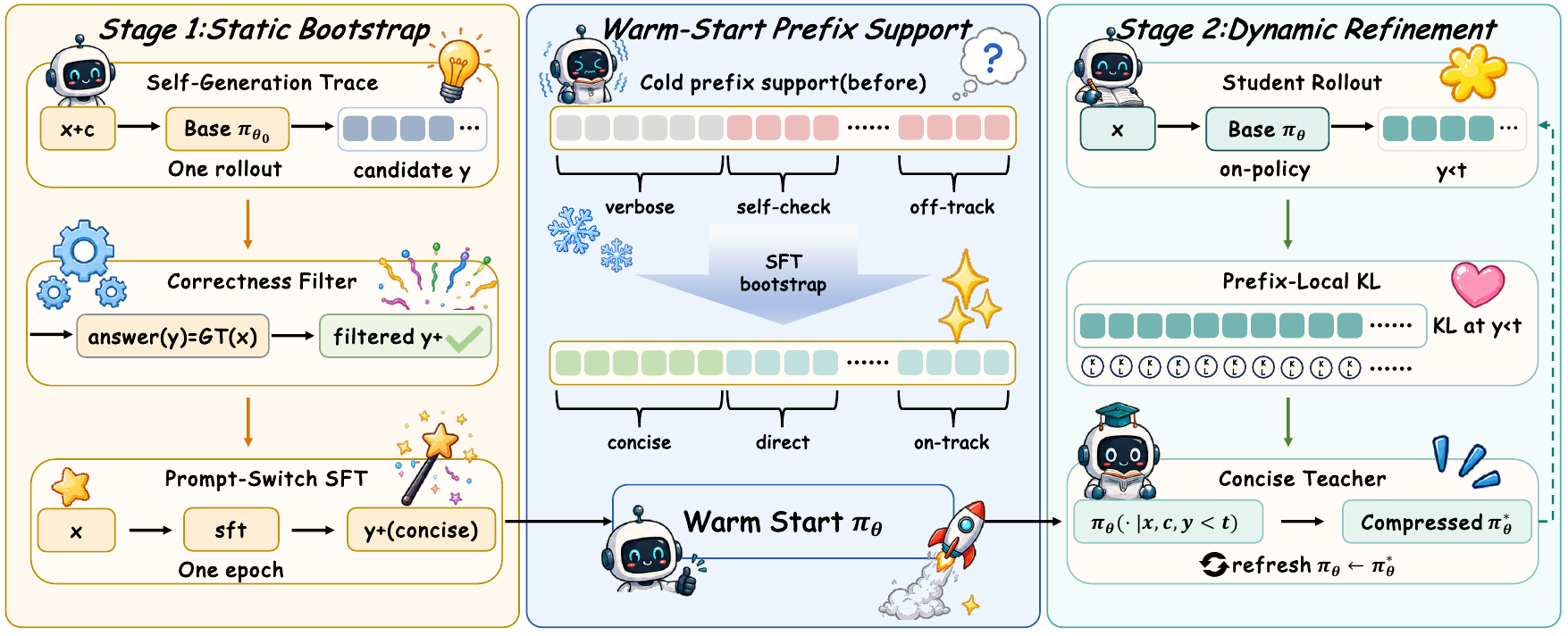}
\caption{Overview of \textsc{BIRD}. Stage~1 constructs a static target from one self-generated concise trace per problem, filters it with ground-truth verification, and applies prompt-switch SFT. The traces are generated with the conciseness instruction $x+c$ but learned under the original prompt $x$. This offline bootstrap moves the student from cold prefix support, which may contain detours and self-checks, to warmer support that is more concise, direct, and on-track. Stage~2 then performs prefix-local reverse-KL self-distillation from this warm start, using the concise teacher only as a dynamic target and no ground-truth answers.}
\label{fig:method}
\end{figure*}

\begin{algorithm}[tb]
\caption{\textsc{BIRD}: Bootstrapped Iterative Self-Reasoning Distillation}
\label{alg:bird}
\textbf{Input}: Base model $\pi_{\theta_0}$, data $\mathcal{D}$, instruction $c$

\textbf{Output}: Compressed model $\pi_{\theta^*}$
\begin{algorithmic}[1]
\State \textbf{Stage 1: bootstrap concise prefix support}
\State $\mathcal{D}_{\mathrm{filtered}}\leftarrow \emptyset$
\For{each $(x_i,\mathrm{GT}_i) \in \mathcal{D}$}
    \State Sample $y_i \sim \pi_{\theta_0}(\cdot \mid x_i,c)$
    \State Extract $\hat a_i=\mathrm{answer}(y_i)$
    \If{$\hat a_i = \mathrm{GT}_i$}
        \State $\tilde y_i \leftarrow \mathrm{Truncate}(y_i,T_{\mathrm{SFT}})$
        \State $\mathcal{D}_{\mathrm{filtered}}\leftarrow
        \mathcal{D}_{\mathrm{filtered}}\cup\{(x_i,\tilde y_i)\}$
    \EndIf
\EndFor
\State $\theta_{\mathrm{SFT}}\leftarrow
\mathrm{LoRA\text{-}SFT}(\theta_0,\mathcal{D}_{\mathrm{filtered}};\,(x_i+c)\rightarrow x_i)$
\State Set $\theta \leftarrow \theta_{\mathrm{SFT}}$
\State \textbf{Stage 2: warm-start on-policy self-distillation}
\State $\bar{\theta} \leftarrow \mathrm{StopGrad}(\theta)$
\For{training step $k=1,2,\dots,K$}
    \If{$k \bmod M = 0$}
        \State $\bar{\theta} \leftarrow \mathrm{StopGrad}(\theta)$
    \EndIf
    \State Sample a batch $\mathcal{B}\subset \mathcal{D}$
    \For{each $x_i \in \mathcal{B}$}
        \State Sample student rollout $y_i \sim \pi_\theta(\cdot \mid x_i)$
        \State $T_i\leftarrow \min(|y_i|,T_{\mathrm{KL}})$
        \State $\ell_{i,t}\leftarrow
        D_{\mathrm{KL}}\!\left(
        \pi_\theta(\cdot \mid x_i,y_{i,<t})
        \,\|\, 
        \pi_{\bar{\theta}}(\cdot \mid x_i,c,y_{i,<t})
        \right)$
        \State $\mathcal{L}_i \leftarrow \sum_{t=1}^{T_i}\ell_{i,t}$
    \EndFor
    \State $\mathcal{L}_{\mathcal{B}}\leftarrow
    |\mathcal{B}|^{-1}\sum_{x_i\in\mathcal{B}}\mathcal{L}_i$
    \State Update $\theta$ with $\nabla_\theta \mathcal{L}_{\mathcal{B}}$
\EndFor
\State $\theta^* \leftarrow \theta$
\Return $\pi_{\theta^*}$
\end{algorithmic}
\end{algorithm}

\subsection{The Cold-Start Problem in OPSD}

OPSD trains on prefixes generated by the current student rather than on a fixed offline dataset. At each training step, the student samples a rollout from the original task prompt, and the training prefixes are taken from this rollout. The same prefixes are then evaluated by both the student and a teacher. The teacher is a stop-gradient copy of the model conditioned on an additional conciseness instruction. It does not receive the ground-truth answer or the student's final answer as input, and provides next-token distributions only at prefixes already produced by the student. The loss compares the student and teacher distributions at each prefix, while gradients are applied only to the student.

In this setup, the student policy $\pi_\theta$ is initialized from the base model $\pi_{\theta_0}$ and optimized with a reverse-KL objective. Let $q_t=\pi_\theta(\cdot\mid x,y_{<t})$ and $p_t=\pi_{\bar{\theta}}(\cdot\mid x,c,y_{<t})$, where $\bar{\theta}$ denotes the teacher parameters implemented as a periodically updated stop-gradient copy of $\theta$. The training objective is:
\begin{equation}
\mathcal{L}(\theta)
=
\mathbf{E}_{x\sim\mathcal D,\; y\sim\pi_\theta(\cdot\mid x)}
\Bigl[\sum_t D_{\mathrm{KL}}(q_t\|p_t)\Bigr].
\label{eq:opsd}
\end{equation}

Both $q_t$ and $p_t$ are defined at the same student-sampled prefix $y_{<t}$, and the rollout $y$ itself is sampled from the current policy. Thus, the support on which KL supervision is applied is determined by the prefixes that the student already visits. When training starts from the base model, many visited prefixes are verbose, redundant, or off-track. The KL objective then matches the concise teacher in trajectory regions that a compressed reasoner should ideally avoid. The issue is therefore not simply how to construct a concise teacher, but how to place the on-policy matching objective on informative prefix support. Poor initialization can make this support inefficient for compression, slowing convergence and limiting the attainable compression frontier.

\subsection{BIRD: Two-Stage Self-Distillation}

The key idea behind \textsc{BIRD}, summarized in Algorithm~\ref{alg:bird} and Figure~\ref{fig:method}, is to decompose distribution matching into an offline bootstrap followed by on-policy refinement. The offline stage first shifts the student's rollout-induced prefix support, while the online stage then applies the same prefix-local OPSD objective on the resulting warm-started support. Thus, the two stages differ less in the form of the optimization objective than in the rollout distribution on which the matching signal is  actually evaluated.

The bootstrap stage uses the base model itself to construct a simple correctness-filtered static target. For each training problem $x$, we sample one concise rollout $y \sim \pi_{\theta_0}(\cdot \mid x,c)$ under the conciseness instruction $c$, and retain the rollout only when its final answer matches the ground truth:
\begin{equation}
\mathcal{D}_{\text{filtered}}
=
\{(x,y)\mid \mathrm{answer}(y)=\mathrm{GT}(x)\}.
\end{equation}

Ground truth is used only for this binary verification step and is never provided as part of the model input  during training. We then perform one epoch of LoRA SFT on $\mathcal{D}_{\text{filtered}}$ by maximizing the likelihood of the retained concise traces under the original task prompt. A central design choice is the prompt switch: traces are generated with the teacher prompt $x+c$, but SFT training uses the student prompt $x$ without the conciseness instruction. The resulting model $\pi_{\theta_{\mathrm{SFT}}}$ is encouraged to internalize instruction-induced conciseness as a default reasoning behavior rather than relying on an inference-time prompt. We keep the bootstrap deliberately lightweight, using a single rollout per problem, a 2,048-token cutoff, one training epoch, and no external teacher model. These choices limit dependence on a fixed offline trace set while still moving the student toward concise and on-track prefixes before on-policy training begins.

Starting from $\pi_{\theta_{\mathrm{SFT}}}$, \textsc{BIRD} performs on-policy reverse-KL self-distillation using Eq.~\ref{eq:opsd}, again with LoRA. The teacher parameters $\bar{\theta}$ are periodically refreshed from the student, and the KL loss is computed on the first 1,024 tokens of each rollout. The objective is the same as in conventional OPSD. The critical difference is the initialization. Optimization begins from $\pi_{\theta_{\mathrm{SFT}}}$ rather than $\pi_{\theta_0}$. Since the warm-started policy already visits prefixes from shorter and more reliable trajectories, the prefix-local KL terms are evaluated on cleaner contexts. This allows on-policy optimization to refine useful reasoning behavior instead of spending much of its signal in verbose or off-track regions inherited from the base model.

\begin{table*}[t]
\centering
\small
\setlength{\tabcolsep}{7pt}
\begin{tabular}{llccccccccc}
\toprule
\multirow{2}{*}{Model} & \multirow{2}{*}{Method} & \multicolumn{3}{c}{MATH-500} & \multicolumn{3}{c}{AIME 2024} & \multicolumn{3}{c}{AIME 2025} \\
\cmidrule(lr){3-5} \cmidrule(lr){6-8} \cmidrule(lr){9-11}
& & Acc $\uparrow$ & Len $\downarrow$ & TE $\uparrow$ & Acc $\uparrow$ & Len $\downarrow$ & TE $\uparrow$ & Acc $\uparrow$ & Len $\downarrow$ & TE $\uparrow$ \\
\midrule
\multirow{4}{*}{\textbf{Qwen3-1.7B}} & Base & 79.40 & 2673 & 10.06 & 25.00 & 5867 & 2.88 & 25.00 & 4677 & 2.96 \\
& Concise & \textbf{82.00} & 1949 & 10.83 & 27.92 & 4937 & 3.28 & 25.42 & 4354 & 3.03 \\
& CRISP & 81.40 & 1515 & 11.11 & \textbf{30.83} & 5251 & \textbf{3.60} & \textbf{25.83} & 4532 & 3.07 \\
& \textbf{BIRD} & 79.60 & \textbf{1092} & \textbf{11.38} & 27.50 & \textbf{3837} & 3.33 & \textbf{25.83} & \textbf{4322} & \textbf{3.09} \\
\midrule
\multirow{4}{*}{\textbf{Qwen3-4B}} & Base & 81.60 & 2767 & 10.30 & 39.17 & 5659 & 4.53 & 29.58 & 5440 & 3.44 \\
& Concise & \textbf{85.40} & 1853 & 11.35 & 42.50 & 4890 & 5.00 & 34.17 & 4805 & 4.03 \\
& CRISP & 82.20 & 1601 & 11.14 & 45.83 & 4680 & 5.42 & \textbf{40.83} & 4594 & 4.84 \\
& \textbf{BIRD} & \textbf{85.40} & \textbf{1034} & \textbf{12.30} & \textbf{49.17} & \textbf{4038} & \textbf{5.92} & \textbf{40.83} & \textbf{4200} & \textbf{4.89} \\
\midrule
\multirow{4}{*}{\textbf{Qwen3-8B}} & Base & 86.20 & 3099 & 10.72 & 38.33 & 6230 & 4.39 & 26.25 & 5595 & 3.04 \\
& Concise & 89.20 & 2045 & 11.70 & 50.00 & 5210 & 5.84 & 34.58 & 4953 & 4.06 \\
& CRISP & 89.60 & 2102 & 11.71 & 53.33 & 4938 & 6.27 & 39.58 & 4577 & 4.70 \\
& \textbf{BIRD} & \textbf{92.00} & \textbf{1115} & \textbf{13.11} & \textbf{58.33} & \textbf{4067} & \textbf{7.02} & \textbf{42.08} & \textbf{3991} & \textbf{5.08} \\
\bottomrule
\end{tabular}
\caption{Main results across three model scales and three benchmarks. TE = accuracy / $\ln$(response length). MATH-500: $n{=}1$ (pass@1); AIME: $n{=}8$ (mean@8). CRISP and BIRD values are taken at the checkpoint with the highest TE on each benchmark.}
\label{tab:main}
\end{table*}
\section{Experiments}
\label{sec:experiments}

\subsection{Experimental Setup}

\paragraph{Datasets.} All training methods are trained on DAPO-Math-17k-dedup \cite{yu2026dapo}. We evaluate on MATH-500, AIME 2024, and AIME 2025. MATH-500 is evaluated with pass@1 ($n=1$), whereas the AIME benchmarks are evaluated with mean@8 ($n=8$) \cite{chen2021evaluating}.

\paragraph{Models and Training.} We evaluate four model configurations spanning the Qwen3 and DeepSeek-R1-Distill-Llama families, including Qwen3-1.7B, Qwen3-4B, Qwen3-8B \cite{yang2025qwen3}, and DeepSeek-R1-Distill-Llama-8B \cite{guo2025deepseek}. All training-based methods use LoRA adaptation. For fair comparison, CRISP and \textsc{BIRD} are trained for one epoch, with checkpoints saved every 25 steps. All experiments are conducted using 8 NVIDIA A800 GPUs.

\paragraph{Metrics and Baselines.} During inference, we set the temperature to 0.6, top-$p$ to 0.95, top-$k$ to 20, and the maximum generation length to 8,192 tokens. Our primary metric is Token Efficiency (TE), defined as:
\begin{equation}
\mathrm{TE}
=
\frac{\mathrm{Accuracy}}
{\ln(\mathrm{Length})},
\label{eq:te}
\end{equation}
where Accuracy denotes the average fraction of correct answers across all sampled solutions, and Length denotes the average response length in tokens, excluding truncated outputs. The logarithmic denominator captures diminishing returns from reducing output length as responses become shorter, thereby rewarding relative rather than absolute compression. Unless otherwise specified, we report the checkpoint that achieves the highest TE on each benchmark. The baselines are \textbf{Base}, the original model without any conciseness instruction, \textbf{Concise}, which applies the conciseness instruction only at inference time without additional training, and \textbf{CRISP}, a cold-start on-policy self-distillation baseline corresponding to our reimplementation of CRISP.

\subsection{Main Results and Scaling}

Table~\ref{tab:main} summarizes the performance of \textsc{BIRD} and competing methods across three Qwen3 model scales and three benchmarks. Overall, \textsc{BIRD} achieves the highest Token Efficiency (TE) in 8 out of 9 model-benchmark pairs, and its advantage over cold-start OPSD becomes more pronounced as model scale increases. At 1.7B, the gains are driven mainly by stronger compression rather than accuracy improvement. For example, compared with CRISP on MATH-500, \textsc{BIRD} reduces response length by 28\% while incurring only a 1.8-point accuracy drop, yielding a net TE gain. At 4B, \textsc{BIRD} starts to improve accuracy and efficiency jointly. The effect is strongest at 8B, where \textsc{BIRD} consistently achieves higher accuracy with shorter responses than CRISP across all three benchmarks. For example, on AIME~2024, it improves accuracy by 5.0 points while using 18\% fewer tokens.

This scaling trend suggests that larger models benefit more from the proposed two-stage procedure. A plausible explanation is that larger models follow the conciseness instruction more reliably during Stage~1, producing cleaner initial prefix support for OPSD, while their greater representational capacity leaves more room for Stage~2 to compress reasoning traces without sacrificing correctness. Notably, applying a conciseness instruction only at inference time, without any additional training, already improves over the original base model, suggesting that concise reasoning behavior is latent in pretrained reasoning models. \textsc{BIRD} turns this instruction-induced behavior into a persistent default and achieves substantially stronger compression than prompting alone.

\begin{table*}[t]
\centering
\small
\setlength{\tabcolsep}{7pt}
\begin{tabular}{lccccccccc}
\toprule
& \multicolumn{3}{c}{MATH-500} & \multicolumn{3}{c}{AIME 2024} & \multicolumn{3}{c}{AIME 2025} \\
\cmidrule(lr){2-4} \cmidrule(lr){5-7} \cmidrule(lr){8-10}
Method & Acc $\uparrow$ & Len $\downarrow$ & TE $\uparrow$ & Acc $\uparrow$ & Len $\downarrow$ & TE $\uparrow$ & Acc $\uparrow$ & Len $\downarrow$ & TE $\uparrow$ \\
\midrule
Base & 73.40 & 2003 & 9.66 & 31.25 & 4990 & 3.67 & 22.92 & 4816 & 2.70 \\
Concise & 73.40 & 1622 & 9.93 & 36.67 & 4573 & 4.35 & 22.92 & \textbf{4081} & 2.76 \\
CRISP & 73.20 & \textbf{1287} & 10.22 & 34.58 & 4826 & 4.08 & 25.83 & 4417 & 3.08 \\
\textbf{BIRD} & \textbf{81.40} & 1342 & \textbf{11.30} & \textbf{41.25} & \textbf{4486} & \textbf{4.91} & \textbf{27.92} & 4197 & \textbf{3.35} \\
\bottomrule
\end{tabular}
\caption{Cross-model generalization on DeepSeek-R1-Distill-Llama-8B, a Llama-based reasoning model whose base responses are already relatively compact. \textsc{BIRD} retains its accuracy-efficiency advantage beyond the Qwen3 family.}
\label{tab:cross}
\end{table*}

\begin{figure}[t]
\centering
\includegraphics[width=0.95\textwidth]{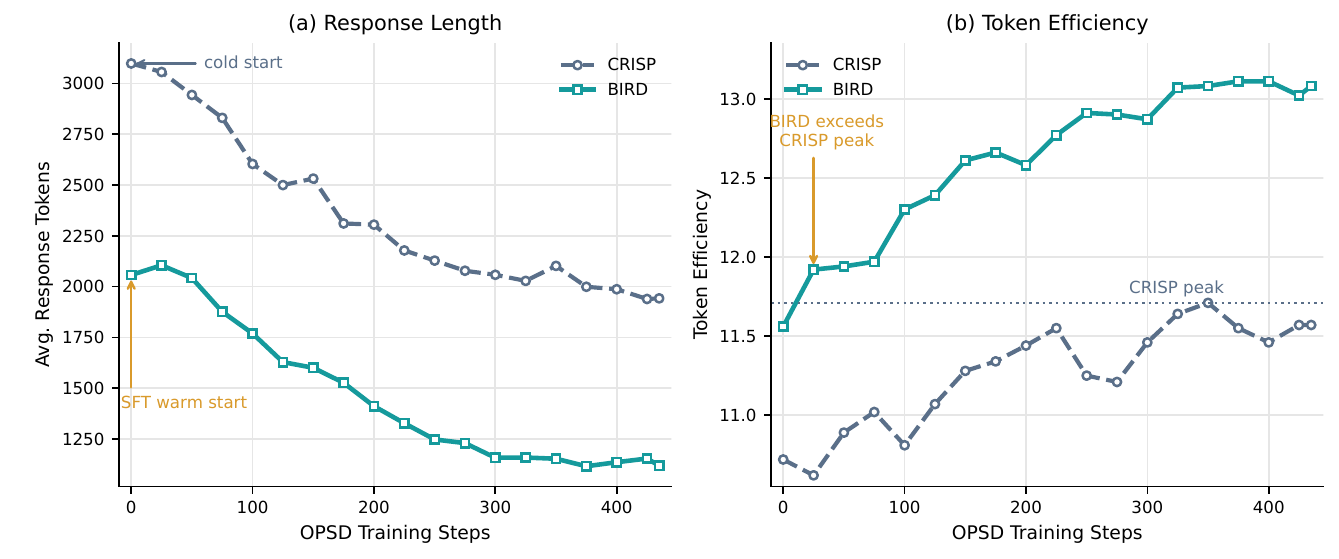}
\caption{Training dynamics of CRISP and \textsc{BIRD} on Qwen3-8B evaluated on MATH-500. Top: average response length. Bottom: Token Efficiency (TE). \textsc{BIRD} starts from the SFT-warmed checkpoint before OPSD updates and maintains lower response length and higher TE throughout training.}
\label{fig:training}
\end{figure}

\subsection{Cross-Model Generalization}

To examine whether \textsc{BIRD} generalizes beyond the Qwen family, we apply the same training recipe to DeepSeek-R1-Distill-Llama-8B, whose base responses are already much shorter than those of Qwen3, leaving less apparent redundancy to remove. Table~\ref{tab:cross} summarizes the results. Despite this compact starting point, \textsc{BIRD} improves MATH-500 accuracy by 8.0 percentage points while simultaneously reducing response length by 33\%. It also improves the accuracy-efficiency trade-off on both AIME benchmarks. CRISP improves TE over the base model, indicating that cold-start OPSD can still remove some redundancy. However, its gains are smaller than those of \textsc{BIRD}. In particular, on MATH-500, CRISP achieves compression at the cost of a slight accuracy drop, whereas \textsc{BIRD} improves both accuracy and TE. These results suggest that SFT-based prefix-support warm starting is not specific to Qwen3 and can transfer to a reasoning model whose initial outputs are already relatively compact.

\subsection{Training Dynamics}

Figure~\ref{fig:training} shows the evolution of response length and Token Efficiency (TE) during OPSD training for Qwen3-8B on MATH-500, which we use because it is the largest benchmark and has the lowest evaluation variance. Before any OPSD update, \textsc{BIRD} already starts from a more favorable prefix distribution. The SFT-warmed model reduces the average response length from 3,099 to 2,057 tokens, corresponding to a 34\% reduction relative to the base model. On this benchmark, its initial TE already matches the peak TE of CRISP, indicating that even lightweight SFT moves the student away from many verbose prefixes on relatively simple problems. This advantage persists throughout training. After only 25 optimization steps, \textsc{BIRD} reaches a TE of 11.92, already exceeding the best value attained by CRISP. In other words, the performance that CRISP reaches after nearly a full epoch is achieved by \textsc{BIRD} almost immediately. Although both methods plateau before the end of training, they converge to markedly different levels. \textsc{BIRD} stabilizes at a TE of 13.02--13.11, whereas CRISP remains in the range of 11.46--11.71. The resulting gap of approximately 1.4 TE points persists after convergence, suggesting that prolonged cold-start OPSD alone does not close the initialization-induced gap.

\begin{figure*}[t]
\centering
{\includegraphics[width=\textwidth]{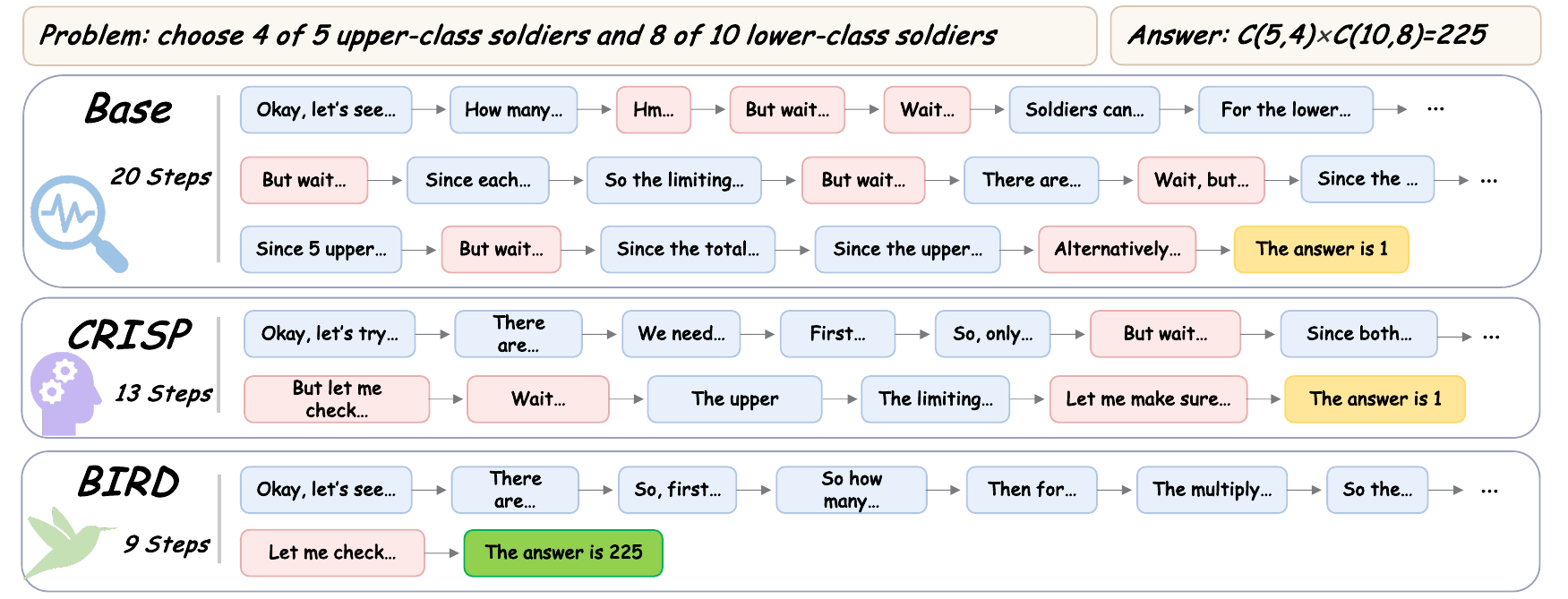}}
\caption{Qualitative trace analysis on a representative MATH-500 counting problem using Qwen3-1.7B. Blue segments denote useful reasoning steps, while red segments denote self-doubt or detours. The Base model produces a long trace with many self-checks and ends incorrectly. CRISP removes some redundancy but retains enough hedging to remain off-track. \textsc{BIRD} follows a direct combinatorial path, reduces self-checks from $8$ to $1$, and reaches the correct answer.}
\label{fig:qual}
\end{figure*}

\subsection{Qualitative Examples}

Figure~\ref{fig:qual} presents reasoning traces from the Base model, CRISP, and \textsc{BIRD} on a representative MATH-500 problem using Qwen3-1.7B. The task requires choosing 4 of 5 upper-class soldiers and 8 of 10 lower-class soldiers, whose direct solution is $\binom{5}{4}\binom{10}{8}=225$. The Base model generates 20 reasoning steps, including 8 redundant self-doubt interruptions, and ultimately produces the incorrect answer 1. CRISP shortens the trace to 13 steps and reduces such interruptions from 8 to 4, but the remaining hedging still derails the solution. In contrast, \textsc{BIRD} requires only 9 steps, contains a single brief self-check, and directly follows the combinatorial counting path to the correct answer. This example shows that excessive self-verification is not only token-inefficient, but can also perturb reasoning by reopening settled subproblems.

\section{Ablation Studies}

\paragraph{Role of each stage.} Table~\ref{tab:abl-stage} compares SFT-only, cold-start OPSD (CRISP), and the full \textsc{BIRD} pipeline on Qwen3-8B. SFT-only already gives a competitive TE, showing that the bootstrap stage can move the model toward shorter and more useful rollouts before any on-policy updates. CRISP also improves over the base model, but it remains below the full pipeline on all three benchmarks. The best results come from applying OPSD after the SFT warm start, suggesting that the two stages are most effective when used together.

\begin{table}[t]
\centering
\begin{tabular}{llccc}
\toprule
Benchmark & Method & Acc $\uparrow$ & Len $\downarrow$ & TE $\uparrow$ \\
\midrule
\multirow{3}{*}{MATH-500} & SFT-only & 88.20 & 2057 & 11.56 \\
& CRISP & 89.60 & 2102 & 11.71 \\
& \textsc{BIRD} & \textbf{92.00} & \textbf{1115} & \textbf{13.11} \\
\midrule
\multirow{3}{*}{AIME 2024} & SFT-only & 50.00 & 5273 & 5.83 \\
& CRISP & 53.33 & 4938 & 6.27 \\
& \textsc{BIRD} & \textbf{58.33} & \textbf{4067} & \textbf{7.02} \\
\midrule
\multirow{3}{*}{AIME 2025} & SFT-only & 34.58 & 4862 & 4.07 \\
& CRISP & 39.58 & 4577 & 4.70 \\
& \textsc{BIRD} & \textbf{42.08} & \textbf{3991} & \textbf{5.08} \\
\bottomrule
\end{tabular}
\caption{Stage necessity ablation on Qwen3-8B. We compare the SFT bootstrap alone, cold-start OPSD (CRISP), and the full \textsc{BIRD} pipeline. The full SFT$\to$OPSD procedure obtains the highest accuracy and TE, as well as the shortest responses.}
\label{tab:abl-stage}
\end{table}

\paragraph{Importance of stage order.} Table~\ref{tab:abl-order} compares the proposed ordering (SFT$\rightarrow$OPSD) with the reversed ordering (OPSD$\rightarrow$SFT) on Qwen3-4B. We first train an OPSD model from the base model, select the checkpoint with the highest average TE across benchmarks, generate concise trajectories from this checkpoint, and then perform SFT. Although the reversed ordering produces shorter responses, it leads to substantial accuracy degradation, particularly on more challenging benchmarks. For example, accuracy on AIME 2025 drops from 40.83\% to 28.33\%. This observation suggests that applying SFT after OPSD may partially overwrite the compression policy learned through on-policy optimization with a fixed offline target, whereas the proposed ordering preserves the adaptability introduced during the second stage.

\begin{table}[t]
\centering
\begin{tabular}{llccc}
\toprule
Benchmark & Stage Order & Acc $\uparrow$ & Len $\downarrow$ & TE $\uparrow$ \\
\midrule
\multirow{2}{*}{MATH-500} & OPSD$\to$SFT & 82.40 & \textbf{873} & 12.17 \\
& SFT$\to$OPSD & \textbf{85.40} & 1034 & \textbf{12.30} \\
\midrule
\multirow{2}{*}{AIME 2024} & OPSD$\to$SFT & 41.25 & \textbf{2926} & 5.17 \\
& SFT$\to$OPSD & \textbf{49.17} & 4038 & \textbf{5.92} \\
\midrule
\multirow{2}{*}{AIME 2025} & OPSD$\to$SFT & 28.33 & \textbf{3384} & 3.49 \\
& SFT$\to$OPSD & \textbf{40.83} & 4200 & \textbf{4.89} \\
\bottomrule
\end{tabular}
\caption{Stage order ablation on Qwen3-4B. We compare applying SFT before OPSD with applying OPSD before SFT. The reversed order produces shorter responses, while the proposed SFT$\to$OPSD order achieves higher accuracy.}
\label{tab:abl-order}
\end{table}

\section{Conclusion}
We presented \textsc{BIRD}, a two-stage distribution-matching method for compressing reasoning traces. The central observation is that on-policy self-distillation is governed not only by the concise teacher, but also by the prefix support on which the teacher is queried. When training starts from a verbose base model, prefix-local KL supervision can be applied to redundant or off-track trajectories, limiting the usefulness of the distillation signal. \textsc{BIRD} addresses this bottleneck by first using a lightweight correctness-filtered SFT bootstrap to shift the student toward shorter and more reliable rollouts, and then applying standard on-policy self-distillation from this warmer initialization. Across model scales, model families, and math reasoning benchmarks, \textsc{BIRD} improves the accuracy-efficiency trade-off over prompting and cold-start OPSD. These results suggest that offline bootstrapping and on-policy refinement provide complementary benefits, and that the placement of supervision over rollout prefixes is a key design factor for efficient reasoning distillation.

\clearpage
\newpage
\bibliographystyle{plainnat}
\setcitestyle{numbers}
\bibliography{ref}

\end{document}